\documentclass[11pt,a4paper]{article}
\usepackage[hyperref]{emnlp2020}
\usepackage{times}
\usepackage{latexsym}

\usepackage{microtype}

\usepackage{graphicx}
\graphicspath{ {./pictures/} }

\usepackage{multirow}

\aclfinalcopy % Uncomment this line for the final submission
\def\aclpaperid{3185} %  Enter the acl Paper ID here

\title{Multi-Instance Multi-Label Learning Networks for Aspect-Category Sentiment Analysis}

\author{
	Yuncong Li \renewcommand{\thefootnote}{\arabic{footnote}}\footnotemark[1] \renewcommand{\thefootnote}{\fnsymbol{footnote}}\footnotemark[1], 
	Cunxiang Yin \renewcommand{\thefootnote}{\arabic{footnote}}\footnotemark[1] \renewcommand{\thefootnote}{\fnsymbol{footnote}}\footnotemark[1], 
	Sheng-hua Zhong \renewcommand{\thefootnote}{\arabic{footnote}}\footnotemark[2] \renewcommand{\thefootnote}{\fnsymbol{footnote}}\footnotemark[2], 
	Xu Pan \renewcommand{\thefootnote}{\arabic{footnote}}\footnotemark[1]\\
	\renewcommand{\thefootnote}{\arabic{footnote}}\footnotemark[1] Baidu Inc., Beijing, China \\
	\texttt{\{liyuncong,yincunxiang,panxu\}@baidu.com} \\
	\renewcommand{\thefootnote}{\arabic{footnote}}\footnotemark[2] College of Computer Science and Software Engineering, Shenzhen University, \\ Shenzhen, China \\
	\texttt{csshzhong@szu.edu.cn} \\
}

\date{}

\begin{document}
\maketitle

\renewcommand{\thefootnote}{\fnsymbol{footnote}}
\footnotetext[1]{Equal contribution}
\footnotetext[2]{Corresponding author}
\renewcommand{\thefootnote}{\arabic{footnote}}

\begin{abstract}
	Aspect-category sentiment analysis (ACSA) aims to predict sentiment polarities of sentences with respect to given aspect categories. To detect the sentiment toward a particular aspect category in a sentence, most previous methods first generate an aspect category-specific sentence representation for the aspect category, then predict the sentiment polarity based on the representation. These methods ignore the fact that the sentiment of an aspect category mentioned in a sentence is an aggregation of the sentiments of the words indicating the aspect category in the sentence, which leads to suboptimal performance. In this paper, we propose a Multi-Instance Multi-Label Learning Network for Aspect-Category sentiment analysis (AC-MIMLLN), which treats sentences as bags, words as instances, and the words indicating an aspect category as the key instances of the aspect category. Given a sentence and the aspect categories mentioned in the sentence, AC-MIMLLN first predicts the sentiments of the instances, then finds the key instances for the aspect categories, finally obtains the sentiments of the sentence toward the aspect categories by aggregating the key instance sentiments. Experimental results on three public datasets demonstrate the effectiveness of AC-MIMLLN \footnote{Data and code are available at https://github.com/l294265421/AC-MIMLLN}.
\end{abstract}

\section{Introduction}

\begin{figure}
	\centering
	\includegraphics[scale=0.2]{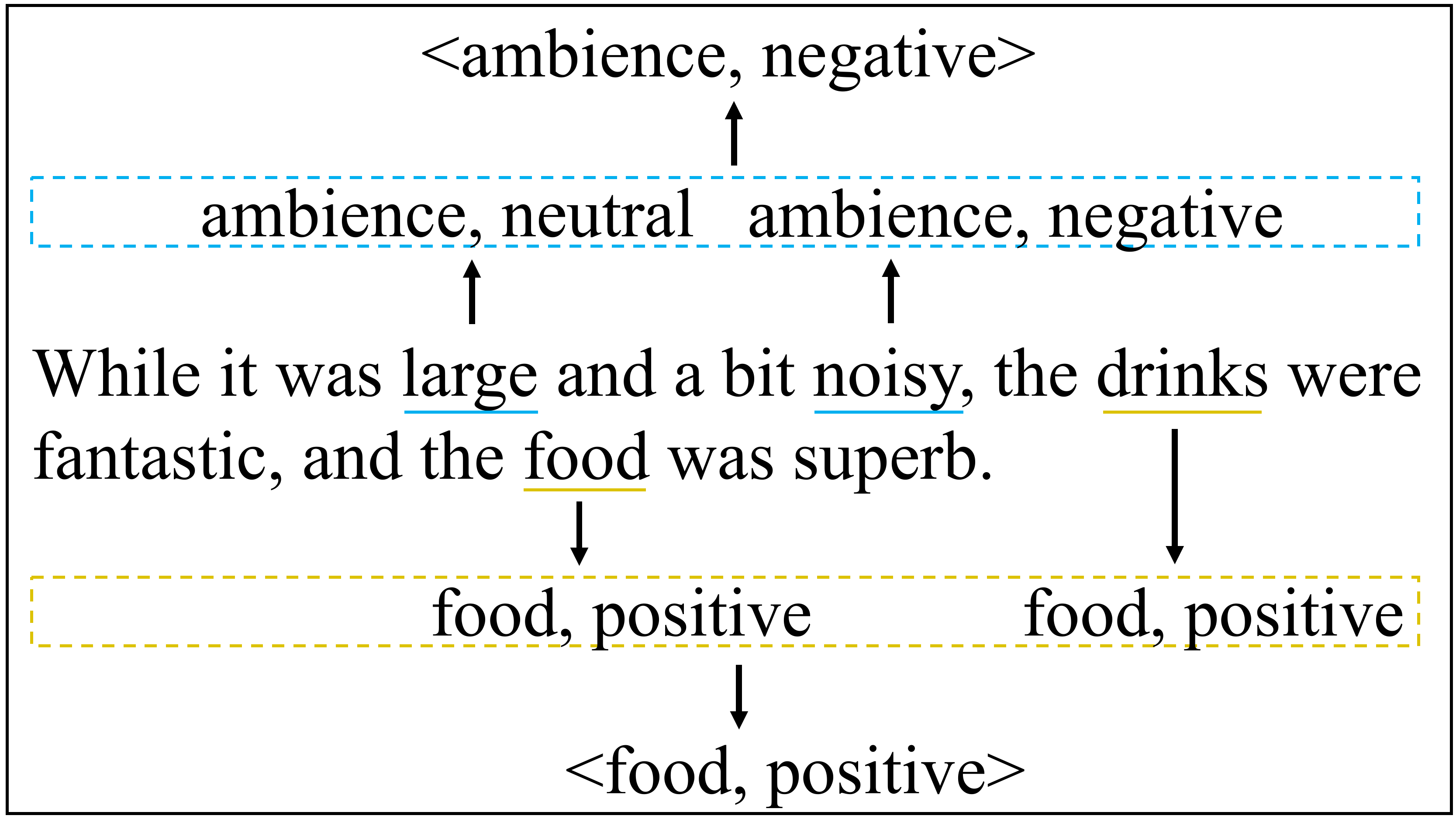}
	\caption{An example of ACD and ACSA. The underlined words are key instances, the labels of the key instances are in the dotted line boxes, and the labels of the sentence are in the angle brackets.}
	\label{fig:examples}
\end{figure}

Sentiment analysis \citep{pang2008opinion,liu2012sentiment} has attracted increasing attention recently. Aspect-based sentiment analysis (ABSA) \citep{pontiki-etal-2014-semeval,pontiki-etal-2015-semeval,pontiki-etal-2016-semeval} is a fine-grained sentiment analysis task and includes many subtasks, two of which are aspect category detection (ACD) that detects the aspect categories mentioned in a sentence and aspect-category sentiment analysis (ACSA) that predicts the sentiment polarities with respect to the detected aspect categories. Figure~\ref{fig:examples} shows an example. ACD detects the two aspect categories, \emph{ambience} and \emph{food}, and ACSA predicts the negative and positive sentiment toward them respectively. In this work, we focus on ACSA, while ACD as an auxiliary task is used to find the words indicating the aspect categories in sentences for ACSA.

Since a sentence usually contains one or more aspect categories, previous studies have developed various methods for generating aspect category-specific sentence representations to detect the sentiment toward a particular aspect category in a sentence. To name a few, attention-based models \citep{wang2016attention,cheng2017aspect,tay2018learning,hu2019can} allocate the appropriate sentiment words for the given aspect category. \citet{xue2018aspect} proposed to generate aspect category-specific representations based on convolutional neural networks and gating mechanisms. Since aspect-related information may already be discarded and aspect-irrelevant information may be retained in an aspect independent encoder, some existing methods \citep{xing2019earlier,liang2019novel} utilized the given aspect to guide the sentence encoding from scratch. Recently, BERT based models \citep{sun2019utilizing,jiang2019challenge} have obtained promising performance on the ACSA task. However, these models ignored that the sentiment of an aspect category mentioned in a sentence is an aggregation of the sentiments of the words indicating the aspect category. It leads to suboptimal performance of these models. For the example in Figure~\ref{fig:examples}, both ``drinks'' and ``food'' indicate the aspect category \emph{food}. The sentiment about \emph{food} is a combination of the sentiments of ``drinks'' and ``food''. Note that, words indicating aspect categories not only contain aspect terms explicitly indicating an aspect category but also contain other words implicitly indicating an aspect category \citep{cheng2017aspect}. In Figure~\ref{fig:examples}, while ``drinks'' and ``food'' are aspect terms explicitly indicating the aspect category \emph{food}, ``large'' and ``noisy'' are not aspect terms implicitly indicating the aspect category \emph{ambience}.

In this paper, we propose a Multi-Instance Multi-label Learning Network for Aspect-Category sentiment analysis (AC-MIMLLN). AC-MIMLLN explicitly models the fact that the sentiment of an aspect category mentioned in a sentence is an aggregation of the sentiments of the words indicating the aspect category. Specifically, AC-MIMLLN treats sentences as bags, words as instances, and the words indicating an aspect category as the key instances \citep{liu2012key} of the aspect category. Given a bag and the aspect categories mentioned in the bag, AC-MIMLLN first predicts the instance sentiments, then finds the key instances for the aspect categories, finally aggregates the sentiments of the key instances to get the bag-level sentiments of the aspect categories.

Our main contributions can be summarized as follows:
\begin{itemize}
	\item We propose a Multi-Instance Multi-Label Learning Network for Aspect-Category sentiment analysis (AC-MIMLLN). AC-MIMLLN explicitly model the process that the sentiment of an aspect category mentioned in a sentence is obtained by aggregating the sentiments of the words indicating the aspect category.
	\item To the best of our knowledge, it is the first time to explore multi-instance multi-label learning in aspect-category sentiment analysis.
	\item Experimental results on three public datasets demonstrate the effectiveness of AC-MIMLLN.
\end{itemize}

\section{Related Work}

\textbf{Aspect-Category Sentiment Analysis} predicts the sentiment polarities with regard to the given aspect categories. Many methods have been developed for this task. \citet{wang2016attention} proposed an attention-based LSTM network, which can concentrate on different parts of a sentence when different aspect categories are taken as input. Some new attention-based methods \citep{cheng2017aspect,tay2018learning,hu2019can} allocated more appropriate sentiment words for aspect categories and obtained bertter performance. \citet{ruder-etal-2016-hierarchical} modeled the interdependencies of sentences in a text with a hierarchical bidirectional LSTM. \citet{xue2018aspect} extracted sentiment features with convolutional neural networks and selectively outputted aspect category related features with gating mechanisms. \citet{xing2019earlier}, \citet{liang2019novel} and \citet{10.1145/3350487} incorporated aspect category information into sentence encoders in the context modeling stage. \citet{lei2019human} proposed a human-like semantic cognition network to simulate the human beings’ reading cognitive process. \citet{sun2019utilizing} constructed an auxiliary sentence from the aspect category and converted ACSA to a sentence-pair classification task. \citet{jiang2019challenge} put forward new capsule networks to model the complicated relationship between aspect categories and contexts. The capsule networks achieved state-of-the-art results. Several joint models \citep{li2017deep,schmitt2018joint,wang2019aspect,Li2019AJM}  were proposed to avoid error propagation, which performed ACD and ACSA jointly. 

However, all these models mentioned above ignored that the sentiment of an aspect category discussed in a sentence is an aggregation of the sentiments of the words indicating the aspect category. 

\textbf{Multi-Instance Multi-Label Learning} (MIMLL) \citep{zhou2006multi} deals with problems where a training example is described by multiple instances and associated with multiple class labels. MIMLL has achieved success in various applications due to its advantages on learning with complicated objects, such as image classification \citep{zhou2006multi,chen2013multi}, text categorization \citep{zhang2008m3miml}, relation extraction \citep{surdeanu2012multi,jiang2016relation}, etc. In ACSA, a sentence contains multiple words (instances) and expresses sentiments to multiple aspect categories (labels), so MIMLL is suitable for ACSA. However, as far as our knowledge, MIMLL has not been explored in ACSA. 

Multiple instance learning (MIL) \citep{keeler1992self} is a special case of MIMLL, where a real-world object described by a number of instances is associated with only one class label. Some studies \citep{kotzias2015group,angelidis-lapata-2018-multiple,pappas2014explaining} have applied MIL to sentiment analysis. \citet{angelidis-lapata-2018-multiple} proposed a Multiple Instance Learning Network (MILNET), where the overarching polarity of a text is an aggregation of sentence or elementary discourse unit polarities, weighted by their importance. An attention-based polarity scoring method is used to obtain the importance of segments. Similar to MILNET, our model also uses an attention mechanism to obtain the importance of instances. However, the attention in our model is learned from the ACD task, while the attention in MILNET is learned from the sentiment classification task. \citet{pappas2014explaining} applied MIL to another subtask of ABSA. They proposed a multiple instance regression (MIR) model to assign sentiment scores to specific aspects of products. However, i) their task is different from ours, and ii) their model is not a neural network.

\section{Model}

In this section, we describe how to apply the multi-instance multi-label learning framework to the aspect-category sentiment analysis task. We first introduce the problem formulation, then describe our proposed Multi-Instance Multi-Label Learning Network for Aspect-Category sentiment analysis (AC-MIMLLN).

\subsection{Problem Formulation}

In the ACSA task, there are $N$ predefined aspect categories $A=\{a_1,a_2,...,a_N\}$ and a predefined set of sentiment polarities $P=\{Neg,Neu,Pos\}$ (i.e., Negative, Neutral and Positive respectively). Given a sentence, $S=\{w_1,w_2,...,w_n\}$ and the $K$ aspect categories, $A^S=\{A^S_1,A^S_2,...,A^S_K\}$, $A^S \subset A$, mentioned in $S$, the ACSA task predicts the sentiment polarity distributions of the $K$ aspect categories, $p=\{p_1,p_2,...,p_K\}$, where $p_k=\{p_{k_{Neg}},p_{k_{Neu}},p_{k_{Pos}}\}$. The multi-instance multi-label learning assumes that, for the $k$-th aspect category, $p_k$ is an unknown function of the unobserved word-level sentiment distributions. AC-MIMLLN first produces a sentiment distribution $p^j$ for each word and then combines these into a sentence-level prediction:
\begin{equation}
p^j=\hat{f}_{\theta_w}(w_j)
\end{equation}
\begin{equation}
p_k=\hat{g}^k_{\theta_S}(p^1,p^2,...,p^n)
\end{equation}

\subsection{Multi-Instance Multi-Label Learning Network for ACSA}

\begin{figure*}
	\centering
	\includegraphics[width=0.99\textwidth]{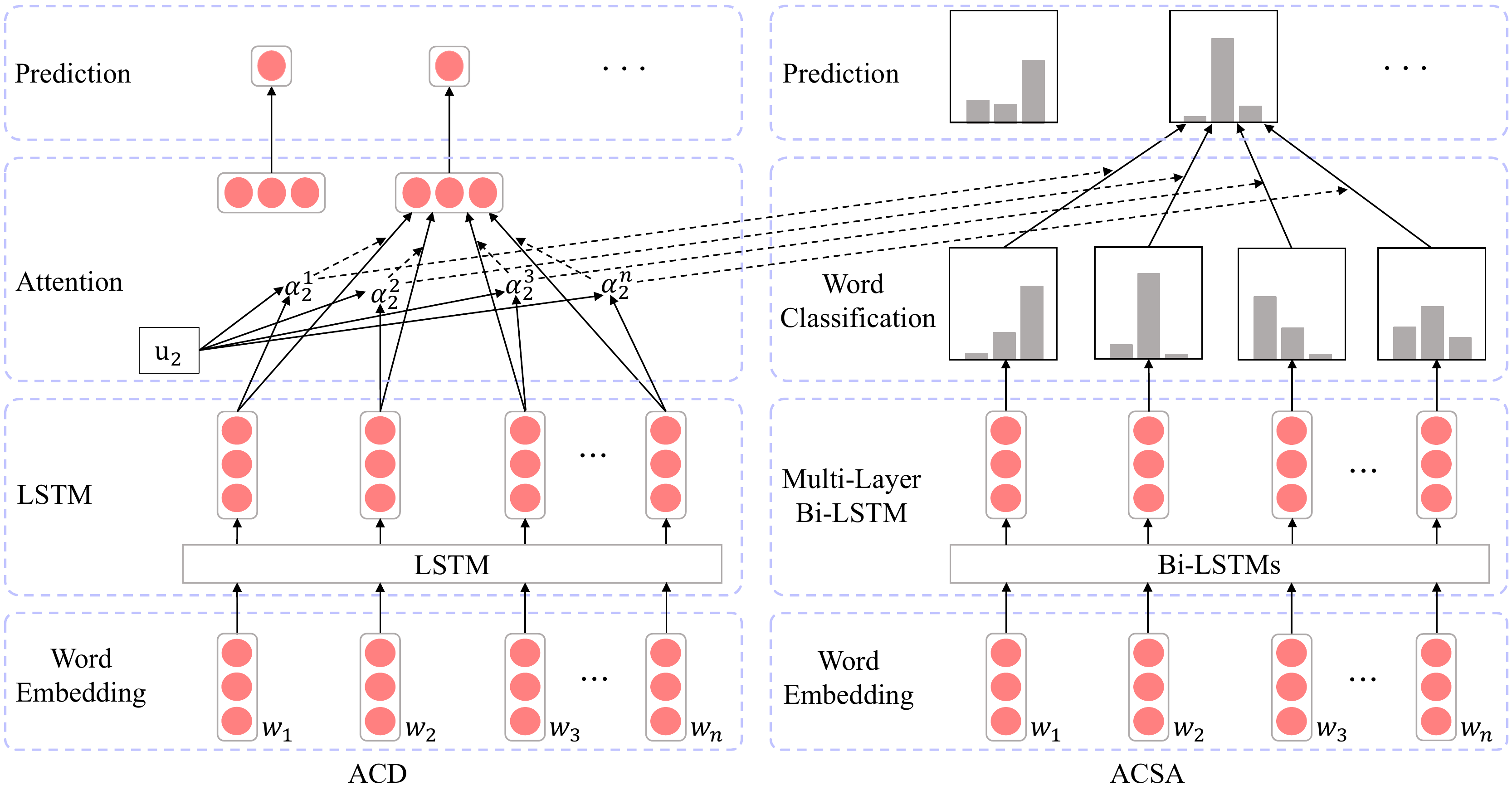}
	\caption{Overall architecture of the proposed method.}
	\label{fig:model}
\end{figure*}

In this section, we introduce our proposed Multi-Instance Multi-Label Learning Network for Aspect-Category sentiment analysis (AC-MIMLLN), which is based on the intuitive assumption that the sentiment of an aspect category mentioned in a sentence is an aggregation of the sentiments of the words indicating the aspect category. In MIMLL, the words indicating an aspect category are called the key instances of the aspect category. Specifically, AC-MIMLLN contains two parts, an attention-based aspect category detection (ACD) classifier and an aspect-category sentiment analysis (ACSA) classifier. Given a sentence, the ACD classifier as an auxiliary task generates the weights of the words for every aspect category. The weights indicate the probabilities of the words being the key instances of aspect categories . The ACSA classifier first predicts the sentiments of the words, then obtains the sentence-level sentiment for each aspect category by combining the corresponding weights and the sentiments of the words. The overall model architecture is illustrated in Figure~\ref{fig:model}. While the ACD part contains four modules: embedding layer, LSTM layer, attention layer and aspect category prediction layer, the ACSA part also consists of four components: embedding layer, multi-layer Bi-LSTM, word sentiment prediction layer and aspect category sentiment prediction layer. In the ACD task, all aspect categories share the embedding layer and the LSTM layer, and have different attention layers and aspect category prediction layers. In the ACSA task, all aspect categories share the embedding layer, the multi-layer Bi-LSTM, and the word sentiment prediction layer, and have different aspect category sentiment prediction layers.

\paragraph{Input:}The input of our model is a sentence consisting of $n$ words $S=\{w_1,w_2,...,w_n\}$.
\paragraph{Embedding Layer for ACD:}The input of this layer is the sentence. With an embedding matrix ${W_w}$, the sentence is converted to a sequence of vectors $X^D=\{x_1^D,x_2^D,...,x_n^D\}$, where,$W_w \in  R^{d\times|V|}$ , $d$ is the dimension of the word embeddings, and $|V|$ is the vocabulary size.
\paragraph{LSTM Layer:}When LSTM \citep{hochreiter1997long} is effective enough, attention mechanisms may not offer effective weight vectors \citep{wiegreffe2019attention}. In order to guarantee the effectiveness of the weights offered by attention mechanisms, we use a single-layer single-direction LSTM for ACD. This LSTM layer takes the word embeddings of the ACD task as input, and outputs hidden states $H=\{h_1,h_2,...,h_n\}$. At each time step $i$, the hidden state $h_i$ is computed by:
\begin{equation}
h_i=LSTM(h_{i-1},x^D_i)
\end{equation}
The size of the hidden state is also set to be $d$.
\paragraph{Attention Layer:}This layer takes the output of the LSTM layer as input, and produce an attention \citep{yang2016hierarchical} weight vector for each predefined aspect category. For the $j$-th aspect category:
\begin{equation}
M_j=tanh(W_jH+b_j),j=1,2,...,N
\end{equation}
\begin{equation}
\alpha_j=softmax(u^T_jM_j),j=1,2,...,N
\end{equation}
where $W_j \in R^{d \times d}$,$b_j \in R^d$,$u_j \in R^d$ are learnable parameters, and $\alpha_j \in R^n$ is the attention weight vector.
\paragraph{Aspect Category Prediction Layer:}We use the weighted hidden state as the sentence representation for ACD prediction. For the $j$-th category:
\begin{equation}
r_j=H\alpha^T_j,j=1,2,...,N
\end{equation}
\begin{equation}
\hat{y}_j=sigmoid(W_jr_j+b_j),j=1,2,...,N
\end{equation}
where $W_j \in R^{d \times 1}$ and $b_j$ is a scalar.

\paragraph{Embedding Layer for ACSA:}For ease of reference, we use different embedding layers for ACD and ACSA. This embedding layer converts the sentence $S$ to a sequence of vectors $X^C=\{x_1^C,x_2^C,...,x_n^C\}$ with the help of the embedding matrix $W_w$. 
\paragraph{Multi-Layer Bi-LSTM:}The output of the embedding layer for ACSA are fed into a multi-layer Bidirectional LSTM \citep{graves2013speech} (Bi-LSTM). Each layer takes the output of the previous layer as input. Formally, given the hidden states of the $(l-1)$-th layer,$H^{l-1}=\{h_1^{l-1},h_2^{l-1},...,h_n^{l-1}\}$, the $l$-th Bi-LSTM outputs hidden states $H^l=\{h_1^l,h_2^l,...,h_n^l\}$. At each time step $i$, the hidden state $h_i^l$ is computed by:
\begin{equation}
\overrightarrow{h^l_i}=\overrightarrow{LSTM}(\overrightarrow{h^l_{i-1}},h^{l-1}_i)
\end{equation}
\begin{equation}
\overleftarrow{h^l_i}=\overleftarrow{LSTM}(\overleftarrow{h^l_{i+1}},h^{l-1}_i)
\end{equation}
\begin{equation}
h^l_i=[\overrightarrow{h^l_i};\overleftarrow{h^l_i}]
\end{equation}
where $H^0=\{x_1^C,x_2^C,...,x_n^C\}$, $\overrightarrow{h_i^l} \in R^{d/2}$,$\overleftarrow{h_i^l} \in R^{d/2}$, $h_i \in R^d$, and $d/2$ denote the size of the hidden state of LSTM. The total number of Bi-LSTM layers is $L$.
\paragraph{Word Sentiment Prediction Layer:}We use the hidden state $h_i^L$ at the time step $i$ of the $L$-th layer Bi-LSTM as the representation of the $i$-th word, and two fully connected layers are used to produce the $i$-th word sentiment prediction $p^i$:
\begin{equation}
p^i=W^2ReLU(W^1h^L_i+b^1)+b^2
\end{equation}
where $W^1 \in R^{d \times d}$, $W^2 \in R^{d \times 3}$, $b^1 \in R^d$, $b^2 \in R^3$ are learnable parameters. Note there is no softmax activation function after the fully connected layer, which lead it difficult to train our model.
\paragraph{Aspect Category Sentiment Prediction Layer:}We obtain the aspect category sentiment predictions by aggregating the word sentiment predictions based on the weights offered by the ACD task. Formally, for the $j$-th aspect category, its sentiment $p_j$ can be computed by:
\begin{equation}
p_j=softmax(\sum_{i=1}^{n}p^i\alpha^i_j)
\end{equation}
where $p_j \in R^3$ , and $\alpha_j^i$ indicates the weight of the $i$-th word about the $j$-th aspect category from the weight vector $\alpha_j$ offered by the ACD task.
\paragraph{Loss:}For the ACD task \footnote{ACD is an auxiliary task. Although AC-MIMLLN performs both ACD and ACSA, the aspect categories it detects (i.e., the results of ACD) are usually ignored in both training stage and testing stage. The reason is that our ACD classifier is simple, it can produce effective attention weights, but may not generate effective predictions for the ACD task. In this paper, we focus on ACSA and only evaluate the performance of AC-MIMLLN on ACSA.}, as each prediction is a binary classification problem, the loss function is defined by:
\begin{equation}
L_A(\theta_A)=-\sum_{j=1}^{N}y_jlog\hat{y}_j+(1-y_j)log(1-\hat{y}_j)
\end{equation}

For the ACSA task, only the loss of the $K$ aspect categories mentioned in the sentence is included, and the loss function is defined by:
\begin{equation}
L_S(\theta_S)=-\sum_{j=1}^{K}\sum_{c \in P}y_{j_c}logp_{j_c}
\end{equation}

We jointly train our model for the two tasks. The parameters in our model are then trained by minimizing the combined loss function:
\begin{equation}
L(\theta)=L_A(\theta_A)+\beta L_S(\theta_S)+\lambda \left \| \theta \right \|^2_2
\end{equation}
where $\beta$ is the weight of ACSA loss, $\lambda$ is the $L2$ regularization factor and $\theta$ contains all parameters of our model.

\section{Experiments}

\subsection{Datasets}

\paragraph{Rest14:}The SemEval-2014 restaurant review (Rest14) \citep{pontiki-etal-2014-semeval} dataset has been widely used. Following previous works \citep{cheng2017aspect,tay2018learning,hu2019can}, we remove samples with conflict polarities. Since there is no official development set for Rest14, we use the split offered by \citet{tay2018learning}. 

\paragraph{Rest14-hard:} Following \citet{xue2018aspect}, we construct Rest14-hard. In Rest14-hard, training set and development set are same as Rest14’s, while test set  is constructed from the test set of Rest14. The test set of Rest14-hard only includes sentences containing at least two aspect categories with different sentiment polaritiess.
\paragraph{MAMS-ACSA:}Since the test set of Rest14-hard is small, we also adopt the Multi-Aspect Multi-Sentiment dataset for Aspect Category Sentiment Analysis (denoted by MAMS-ACSA). MAMS-ACSA is released by \citet{jiang2019challenge}, all sentences in which contain multiple aspect categories with different sentiment polarities.

\begin{table}
	\centering
	\begin{tabular}{ |c|c|c|c|c| } 
		\hline
		\multicolumn{2}{|c|}{Dataset} & Pos. & Neg. & Neu. \\ 
		\hline
		\multirow{3}*{Rest14} & Train & 1855 & 733 & 430 \\ 
		\cline{2-5}
		& Dev & 324 & 106 & 70 \\ 
		\cline{2-5}
		& Test & 657 & 222 & 94 \\ 
		\hline
		Rest14-hard & Test & 21 & 20 & 12 \\ 
		\hline
		\multirow{3}*{MAMS-ACSA}  & Train & 1929 & 2084 & 3077 \\ 
		\cline{2-5}
		& Dev & 241 & 259 & 388 \\ 
		\cline{2-5}
		& Test & 245 & 263 & 393 \\ 
		\hline
	\end{tabular}
	\caption{Statistics of the datasets.}
	\label{datasets}
\end{table}

We select Rest14-hard and MAMS-ACSA that we call hard datasets because most sentences in Rest14 contain only one aspect or multiple aspects with the same sentiment polarity, which makes ACSA degenerate to sentence-level sentiment analysis \citep{jiang2019challenge}.
 Rest14-hard and MAMS-ACSA can measure the ability of a model to detect multiple different sentiment polarities in one sentence toward different aspect categories. Statistics of these three datasets are given in Table~\ref{datasets}.

\begin{table*}
	\centering
	\begin{tabular}{ |l|l|l|l| } 
		\hline
		Methods & Rest14 & Rest14-hard & MAMS-ACSA\\
		\hline
		GCAE \citep{xue2018aspect} & 81.336($\pm$0.883) & 54.717($\pm$4.920) & 72.098$\dagger$\\
		As-capsule \citep{wang2019aspect} & \textbf{82.179($\pm$0.414)} & 60.755($\pm$2.773) & 75.116($\pm$0.473)\\
		CapsNet \citep{jiang2019challenge} & 81.172($\pm$0.631) & 53.962($\pm$0.924) & 73.986$\dagger$\\
		AC-MIMLLN (ours) & 81.603($\pm$0.715) & \textbf{65.283($\pm$2.264)} & \textbf{76.427($\pm$0.704)}\\
		AC-MIMLLN – w/o mil (ours) & 80.596($\pm$0.816) & 64.528($\pm$2.201) & 75.650($\pm$1.100)\\
		AC-MIMLLN-Affine (ours) & 80.843($\pm$0.760) & 64.151($\pm$3.375) & 74.517($\pm$1.299)\\
		\hline
		BERT \citep{jiang2019challenge} & 87.482($\pm$0.906) & 67.547($\pm$5.894) & 78.292$\dagger$\\
		BERT-pair-QA-B \citep{sun2019utilizing} & 87.523($\pm$1.175) & 69.433($\pm$4.368) & 79.134($\pm$0.973)\\
		CapsNet-BERT \citep{jiang2019challenge} & 86.557($\pm$0.943) & 51.321($\pm$1.412) & 79.461$\dagger$\\
		AC-MIMLLN-BERT (ours) & \textbf{89.250($\pm$0.720)} & \textbf{74.717($\pm$3.290)} & \textbf{81.198($\pm$0.606)}\\
		\hline
	\end{tabular}
	\caption{Results of the ACSA task in terms of accuracy (\%, mean$\pm$(std)). $\dagger$ refers to citing from \citet{jiang2019challenge}.}
	\label{experimental-results}
\end{table*}

\begin{table}
	\centering
	\begin{tabular}{ |l|l|l|l|l|l| } 
		\hline
		\multirow{2}{0.9em}{Methods} & \multirow{2}{0.9em}{food} & \multirow{2}{0.9em}{ser\-vice} & \multirow{2}{0.9em}{amb\-ience} & \multirow{2}{0.9em}{price} & \multirow{2}{0.9em}{misc}\\
		&  &  &  &  & \\
		\hline
		As-capsule & 82.7 & 90.1 & \textbf{84.3} & 80.5 & \textbf{74.6}\\
		\hline
		\multirow{2}{0.9em}{AC-MIMLLN} & \multirow{2}{0.9em}{\textbf{83.7}} & \multirow{2}{0.9em}{\textbf{90.5}} & \multirow{2}{0.9em}{83.6} & \multirow{2}{0.9em}{\textbf{84.0}} & \multirow{2}{0.9em}{69.0}\\
		&  &  &  &  & \\
		\hline
	\end{tabular}
	\caption{Results of the ACSA task on Rest14's aspect categories in terms of accuracy (\%).}
	\label{performances-on-different-aspect-categories}
\end{table}

\subsection{Comparison Methods}
We compare AC-MIMLLN with various baselines. (1) non-BERT models: GCAE \citep{xue2018aspect}, As-capsule \citep{wang2019aspect} \footnote{As-capsule is also a multi-task model, which performs ACD and ACSA simultaneously like our model.} and CapsNet \citep{jiang2019challenge}; (2) BERT \citep{devlin2019bert} based models: BERT \citep{jiang2019challenge}, BERT-pair-QA-B \citep{sun2019utilizing} and CapsNet-BERT \citep{jiang2019challenge}. We also provide the comparisons of several variants of AC-MIMLLN:

\textbf{AC-MIMLLN – w/o mil} generates aspect category-specific representations for the ACAC task. The representations are the weighted sum of the word representations based on the weights offered by the ACD task.

\textbf{AC-MIMLLN-Affine} replaces the LSTM in AC-MIMLLN with an affine hidden layer, which is used to evaluate the effectiveness of the attention in AC-MIMLLN \citep{wiegreffe2019attention}.

\textbf{AC-MIMLLN-BERT} replaces the embedding layer for ACSA and the multi-layer Bi-LSTM in AC-MIMLLN with the uncased basic pre-trained BERT. Since the overall sentiment of a sentence as context information is important for infering the sentiment of a particular aspect category, AC-MIMLLN-BERT also predicts the sentiment of the token ``[CLS]'' and assigns weight 1 to it. AC-MIMLLN-BERT takes ``[CLS] sentence [SEP] aspect category [SEP]'' as input like CapsNet-BERT.

\subsection{Implementation Details}
We implement our models in PyTorch \citep{paszke2017automatic}. We use 300-dimentional word vectors pre-trained by GloVe \citep{pennington-etal-2014-glove} to initialize the word embedding vectors. The batch sizes are set to 32 and 64 for non-BERT models on the Rest14(-hard) dataset and the MAMS-ACSA dataset, respectively, and 16 for BERT-based models. All models are optimized by the Adam optimizer \citep{kingma2014adam}. The learning rates are set to 0.001 and 0.00002 for non-BERT models and BERT-based models, respectively. We set $L=3$, $\lambda=0.00001$ and $\beta=1$. For the ACSA task, we apply a dropout of $p=0.5$ after the embedding and Bi-LSTM layers. For AC-MIMLLN-BERT, ACD is trained first then both of ACD and ACSA are trained together. For other models, ACD and ACSA are directly trained jointly. We apply early stopping in training and the patience is 10. We run all models for 5 times and report the average results on the test datasets.

\begin{table*}
	\centering
	\begin{tabular}{ |l|l|l|l|l|l|l| } 
		\hline
		Methods & \multicolumn{2}{|c|}{Rest14} & \multicolumn{2}{|c|}{Rest14-hard} & \multicolumn{2}{|c|}{MAMS-ACSA}\\
		\hline
		& KID($F_1$) & KISC(acc) & KID($F_1$) & KISC(acc) & KID($F_1$) & KISC(acc)\\
		\hline
		AC-MIMLLN & 38.132 & 73.731 & 48.927 & 62.250 & 69.480 & 68.804\\
		\hline
		AC-MIMLLN-Affine & 60.035 & 76.503 & 66.920 & 67.250 & \textbf{75.083} & 68.503\\
		\hline
		AC-MIMLLN-BERT & \textbf{63.477} & \textbf{83.894} & \textbf{70.027} & \textbf{72.250} & 74.172 & \textbf{75.666}\\
		\hline
	\end{tabular}
	\caption{Performance of detecting the key instances (KID) of the given aspect category in terms of accuracy (\%) and classifying the sentiments of the given key instances (KISC) in terms of $F_1$ measure (\%).}
	\label{performance-on-instance-level}
\end{table*}

\subsection{Experimental Results}
Experimental results are illustrated in Table~\ref{experimental-results}. According to the experimental results, we can come to the following conclusions. First, AC-MIMLLN outperforms all non-BERT baselines on the Rest14-hard dataset and the MAMS-ACSA dataset, which indicates that AC-MIMLLN has better ability to detect multiple different sentiment polarities in one sentence toward different aspect categories. Second, AC-MIMLLN obtains +1.0\% higher accuracy than AC-MIMLLN – w/o mil on the Rest14 dataset, +0.8\% higher accuracy on the Rest14-hard dataset and +0.8\% higher accuracy on the MAMS-ACSA dataset, which shows that the Multiple Instance Learning (MIL) framework is more suitable for the ACSA task. Third, AC-MIMLLN-BERT surpasses all BERT-based models on all three datasets, indicating that AC-MIMLLN can achieve better performance by using more powerful sentence encoders for ACSA. In addition, AC-MIMLLN can't outperform As-capsule on Rest14. The main reason is that AC-MIMLLN has poor perfmance on the aspect category \emph{misc} (the abbreviation for \emph{anecdotes/miscellaneous}) (see Table~\ref{performances-on-different-aspect-categories} and Figure~\ref{fig:cases} (f)).

\begin{table}
	\centering
	\begin{tabular}{ |l|l|l|l| } 
		\hline
		\multirow{2}{2em}{Methods} & \multirow{2}{2em}{Rest14} & \multirow{2}{2em}{Rest14-hard} & \multirow{2}{2em}{MAMS-ACSA}\\
		&  &  & \\
		\hline
		single-pipeline & 81.459 & 61.509 & 72.231\\
		\hline
		single-joint & 80.329 & 62.641 & 75.605\\
		\hline
		multi-pipeline & \textbf{82.117} & 63.396 & 72.675\\
		\hline
		multi-joint & 81.603 & \textbf{65.283} & \textbf{76.427}\\
		\hline
	\end{tabular}
	\caption{Results of AC-MIMLLN in different multi-task settings on ACSA in terms of accuracy(\%).}
	\label{different-multi-task-settings}
\end{table}

\subsection{Impact of Multi-Task Learning}
AC-MIMLLN is a multi-task model, which performs ACD and ACSA simultaneously. Multi-task learning \citep{caruana1997multitask} achieves improved performance by exploiting commonalities and differences across tasks.  In this section, we explore the performance of AC-MIMLLN in different multi-task settings on the ACSA task. Specifically, we explore four settings: single-pipeline, single-joint, multi-pipeline and multi-joint. The “single” means that the ACSA task predicts the sentiment of one aspect category in sentences every time, while the “multi” means that the ACSA task predicts the sentiments of all aspect categories in sentences every time. The “pipeline” indicates that ACD is trained first, then ACSA is trained, while the “joint” indicates ACD and ACSA are trained jointly. The multi-joint is AC-MIMLLN.

Experimental results are shown in Table~\ref{different-multi-task-settings}. First, we observe that, multi-* outperform all their counterparts, indicating modeling all aspect categories in sentences simultaneously can improve the performance of the ACSA task. Second, *-joint surpass *-pipeline on the Rest14-hard dataset and the MAMS-ACSA dataset, which shows that training ACD and ACSA jointly can improve the perfomance on hard datasets. Third, *-joint obtain worse perfomance on the Rest14 dataset than *-pipeline. One possible reason is that  Rest14 is simple and *-joint have bigger model capacity than  *-pipeline and overfit on Rest14.

\subsection{Impact of Multi-layer Bi-LSTM Depth}
\begin{figure}
	\centering
	\includegraphics[scale=0.25]{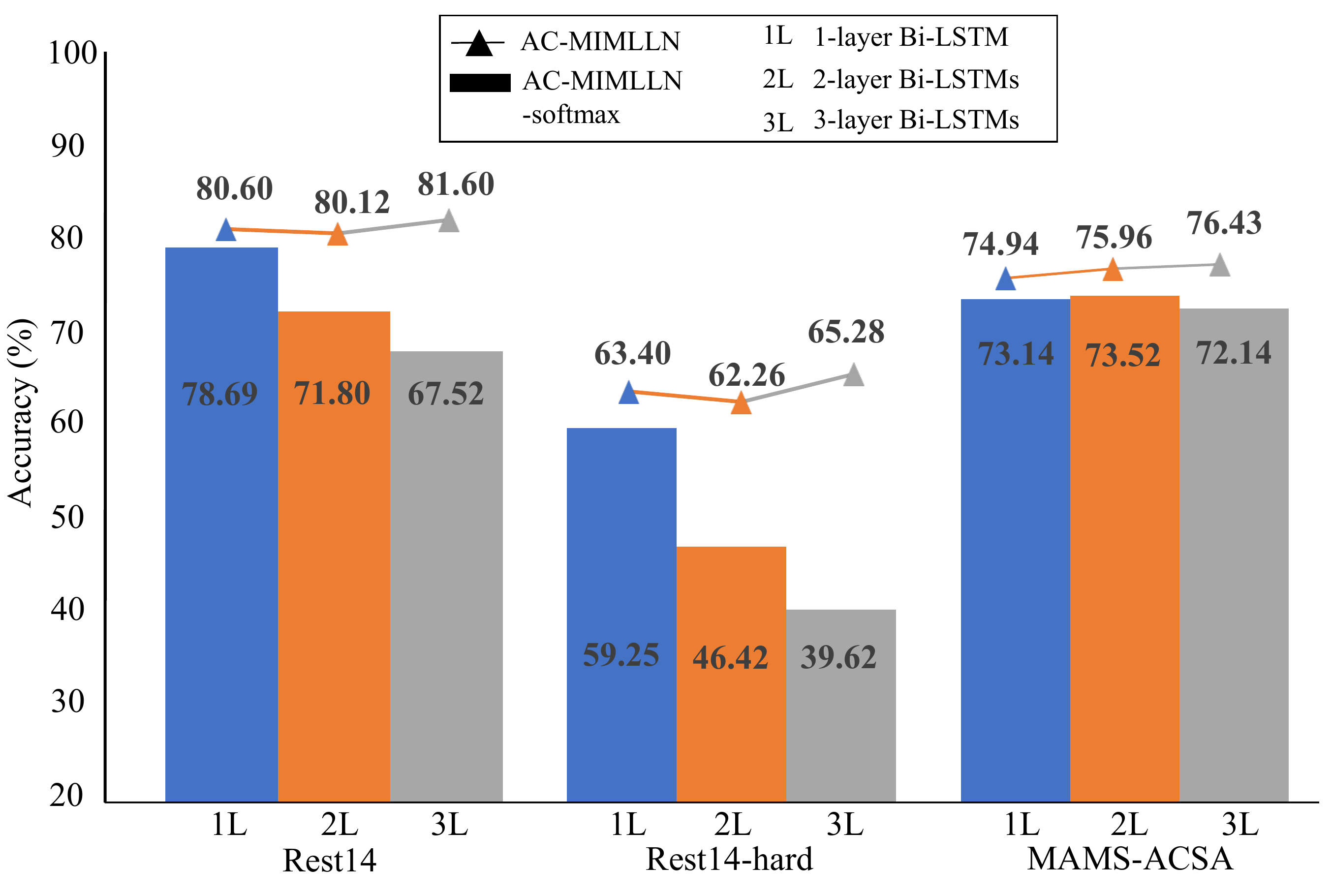}
	\caption{The impact of the number of Bi-LSTM layers and the softmax activation function.}
	\label{fig:softmax}
\end{figure}

In this section, we explore the effect of the number of the Bi-LSTM layers. Experiments results are shown in Figure~\ref{fig:softmax}, which also contains the results of AC-MIMLLN-softmax. AC-MIMLLN-softmax is obtained by adding the softmax activation function to the word sentiment prediction layer of AC-MIMLLN. We observe that, when the number of Bi-LSTM layer increases, AC-MIMLLN usually obtains better performance, and AC-MIMLLN-softmax obtains worse results. It indicates that AC-MIMLLN-softmax is hard to train when its complexity increases, while AC-MIMLLN can achieve better performance by using more powerful sentence encoders for ACSA.

\begin{figure*}
	\centering
	\includegraphics[width=0.99\textwidth]{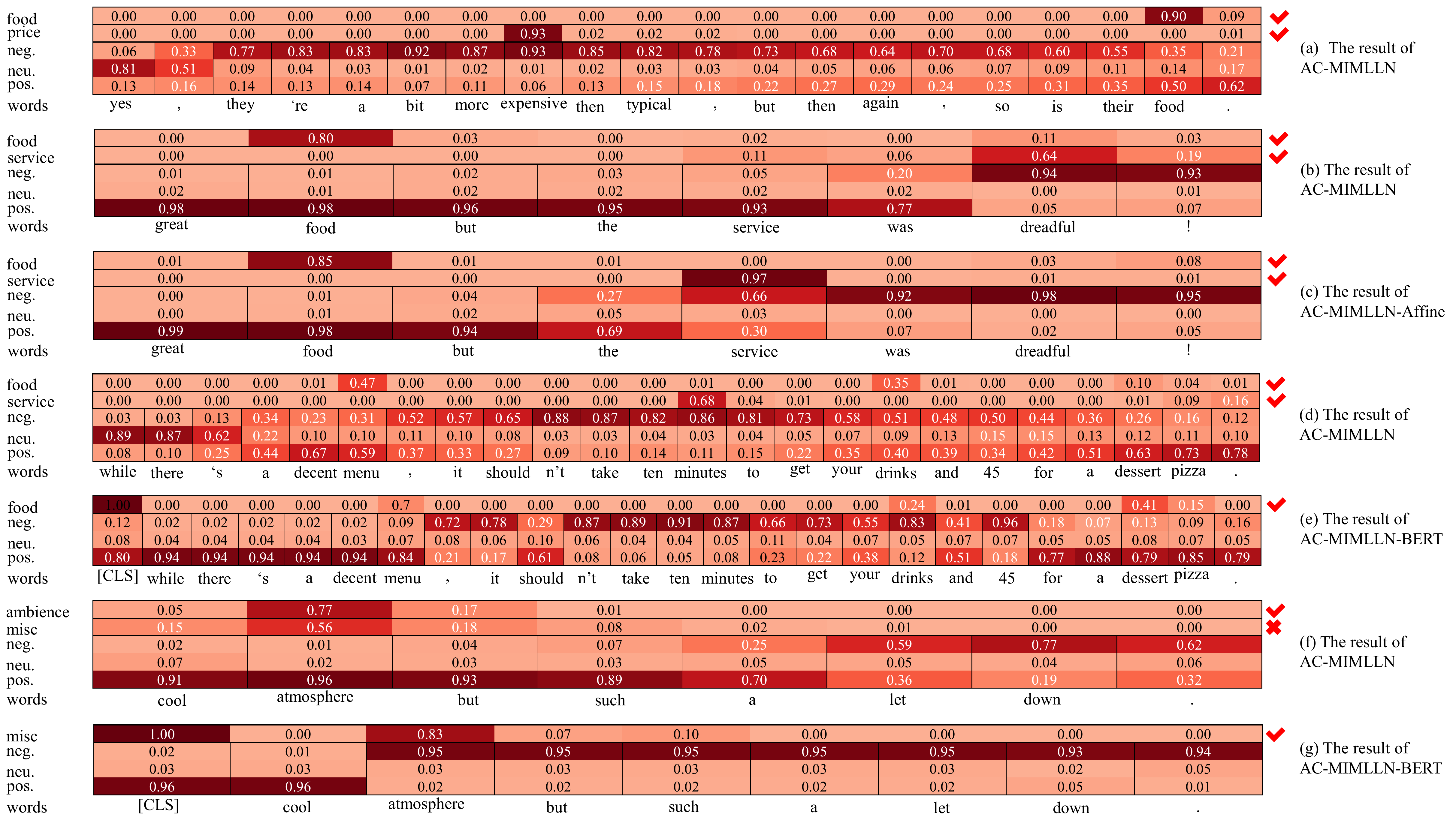}
	\caption{Visualization of the attention weights and the word sentiment prediction results. For each subfigure, the lines corresponding aspect categories show the attention weights offered by the ACD task, while the other three lines show the word sentiment distributions predicted by the ACSA task.}
	\label{fig:cases}
\end{figure*}

\subsection{Quality Analysis}
In this subsection, we show the advantages of our model and analyze where the error lies in through some typical examples and estimating the performance of our model detecting the key instances (KID) of the given aspect category and classifying the sentiments of the given key instances (KISC). We annotate the key instances for the aspect categories mentioned in sentences and their sentiment polarities on the test set of the three datasets. Models judge a word as a key instance if the weight of the word is greater than or equal to 0.1. Experimental results are illustrated in Table~\ref{performance-on-instance-level}. 

\textbf{Case Study} Figure~\ref{fig:cases} visualizes the attention weights and the word sentiment prediction results of four sentences. Figure~\ref{fig:cases} (a) shows that, our model accurately finds the key instances “expensive” for the aspect category \emph{price} and “food” for \emph{food}, and assigns correct sentiments to both the aspect categories and the key instances. Compared with previous models, which generate aspect category-specific sentence representations for the ACSA task directly (e.g. BERT-pair-QA-B) or based on aspect category-related sentiment words (e.g. As-capsule), our model is more interpretable.

In Figure~\ref{fig:cases}, (b) and (c) show that, both AC-MIMLLN and AC-MIMLLN-Affine can correctly predict the sentiments of the aspect categories, \emph{food} and \emph{service}. While AC-MIMLLN-Affine accurately find the key instance “service” for \emph{service}, AC-MIMLLN assigns weights to all the words in the text snippet ``service was dreadful!''. This is because the LSTM-based ACD model in AC-MIMLLN can select useful words for both ACD and ACSA based on the context, which results in better performance (see Table~\ref{experimental-results}). This also can explain why AC-MIMLLN has worse performance on detecting the key instances of the given aspect category than AC-MIMLLN-Affine (see Table~\ref{performance-on-instance-level}).

\textbf{Error Analysis} In Figure~\ref{fig:cases} (d), the sentiments toward “drinks” and “dessert” (key instances of the aspect category \emph{food}) should be neutral, however AC-MIMLLN assigns negative sentiment to "drinks" and positive sentiment to "dessert". Figure~\ref{fig:cases} (e) shows AC-MIMLLN-BERT also assigns wrong sentiments to “drinks” and “dessert”. Table~\ref{performance-on-instance-level} shows that although AC-MIMLLN-BERT significantly improve the performance of KISC, it's results are also less than 80\% on the Rest14-hard dataset and the MAMS-ACSA dataset.

In Figure~\ref{fig:cases} (f), AC-MIMLLN wrongly predict the sentiment of the aspect category \emph{misc}, because it finds the wrong key instances for \emph{misc}. Compared to other aspect categories, it's harder to decide which words are the key instances of \emph{misc} for AC-MIMLLN, resulting in poor performance of AC-MIMLLN on the aspect category \emph{misc}. Figure~\ref{fig:cases} (g) shows AC-MIMLLN-BERT correctly predict the sentiments of the aspect category \emph{misc}, but also finds the wrong key instances for \emph{misc}. Table~\ref{performance-on-instance-level} shows that all results on KID are less than 75\%.

\section{Conclusion}
In this paper, we propose a Multi-Instance Multi-Label Learning Network for Aspect-Category sentiment analysis (AC-MIMLLN). AC-MIMLLN predicts the sentiment of an aspect category mentioned in a sentence by aggregating the sentiments of the words indicating the aspect category in the sentence. Experimental results demonstrate the effectiveness of AC-MIMLLN. Since AC-MIMLLN finds the key instances for the given aspect category and predicts the sentiments of the key instances, it is more interpretable. In some sentences, phrases or clauses rather than words indicate the given aspect category, future work could consider multi-grained instances, including  words, phrases and clauses. Since directly finding the key instances for some aspect categories is ineffective, we will try to first recognize all opinion snippets in a sentence, then assign these snippets to the aspect categories mentioned in the sentence.

\bibliography{anthology,mimlln}

\begin{thebibliography}{40}
\expandafter\ifx\csname natexlab\endcsname\relax\def\natexlab#1{#1}\fi

\bibitem[{Angelidis and Lapata(2018)}]{angelidis-lapata-2018-multiple}
Stefanos Angelidis and Mirella Lapata. 2018.
\newblock \href {https://doi.org/10.1162/tacl_a_00002} {Multiple instance
  learning networks for fine-grained sentiment analysis}.
\newblock \emph{Transactions of the Association for Computational Linguistics},
  6:17--31.

\bibitem[{Caruana(1997)}]{caruana1997multitask}
Rich Caruana. 1997.
\newblock \href {https://doi.org/https://doi.org/10.1023/A:100737960}
  {Multitask learning}.
\newblock \emph{Machine learning}, 28(1):41--75.

\bibitem[{Chen et~al.(2013)Chen, Chi, Fu, and Feng}]{chen2013multi}
Zenghai Chen, Zheru Chi, Hong Fu, and Dagan Feng. 2013.
\newblock \href {https://doi.org/https://doi.org/10.1016/j.neucom.2012.08.001}
  {Multi-instance multi-label image classification: A neural approach}.
\newblock \emph{Neurocomputing}, 99:298--306.

\bibitem[{Cheng et~al.(2017)Cheng, Zhao, Zhang, King, Zhang, and
  Wang}]{cheng2017aspect}
Jiajun Cheng, Shenglin Zhao, Jiani Zhang, Irwin King, Xin Zhang, and Hui Wang.
  2017.
\newblock \href {https://doi.org/https://doi.org/10.1145/3132847.3133037}
  {Aspect-level sentiment classification with heat (hierarchical attention)
  network}.
\newblock In \emph{Proceedings of the 2017 ACM on Conference on Information and
  Knowledge Management}, pages 97--106.

\bibitem[{Devlin et~al.(2019)Devlin, Chang, Lee, and
  Toutanova}]{devlin2019bert}
Jacob Devlin, Ming-Wei Chang, Kenton Lee, and Kristina Toutanova. 2019.
\newblock \href {https://doi.org/https://doi.org/10.18653/v1/N19-1423} {Bert:
  Pre-training of deep bidirectional transformers for language understanding}.
\newblock In \emph{Proceedings of the 2019 Conference of the North American
  Chapter of the Association for Computational Linguistics: Human Language
  Technologies, Volume 1 (Long and Short Papers)}, pages 4171--4186.

\bibitem[{Graves et~al.(2013)Graves, Mohamed, and Hinton}]{graves2013speech}
Alex Graves, Abdel-rahman Mohamed, and Geoffrey Hinton. 2013.
\newblock \href {https://doi.org/https://doi.org/10.1109/ICASSP.2013.6638947}
  {Speech recognition with deep recurrent neural networks}.
\newblock In \emph{2013 IEEE international conference on acoustics, speech and
  signal processing}, pages 6645--6649. IEEE.

\bibitem[{Hochreiter and Schmidhuber(1997)}]{hochreiter1997long}
Sepp Hochreiter and J{\"u}rgen Schmidhuber. 1997.
\newblock \href {https://doi.org/https://doi.org/10.1162/neco.1997.9.8.1735}
  {Long short-term memory}.
\newblock \emph{Neural computation}, 9(8):1735--1780.

\bibitem[{Hu et~al.(2019)Hu, Zhao, Zhang, Cai, Su, Cheng, and Shen}]{hu2019can}
Mengting Hu, Shiwan Zhao, Li~Zhang, Keke Cai, Zhong Su, Renhong Cheng, and
  Xiaowei Shen. 2019.
\newblock \href {https://doi.org/https://doi.org/10.18653/v1/D19-1467} {Can:
  Constrained attention networks for multi-aspect sentiment analysis}.
\newblock In \emph{Proceedings of the 2019 Conference on Empirical Methods in
  Natural Language Processing and the 9th International Joint Conference on
  Natural Language Processing (EMNLP-IJCNLP)}, pages 4593--4602.

\bibitem[{Jiang et~al.(2019)Jiang, Chen, Xu, Ao, and Yang}]{jiang2019challenge}
Qingnan Jiang, Lei Chen, Ruifeng Xu, Xiang Ao, and Min Yang. 2019.
\newblock \href {https://doi.org/https://doi.org/10.18653/v1/D19-1654} {A
  challenge dataset and effective models for aspect-based sentiment analysis}.
\newblock In \emph{Proceedings of the 2019 Conference on Empirical Methods in
  Natural Language Processing and the 9th International Joint Conference on
  Natural Language Processing (EMNLP-IJCNLP)}, pages 6281--6286.

\bibitem[{Jiang et~al.(2016)Jiang, Wang, Li, and Wang}]{jiang2016relation}
Xiaotian Jiang, Quan Wang, Peng Li, and Bin Wang. 2016.
\newblock \href {https://www.aclweb.org/anthology/C16-1139} {Relation
  extraction with multi-instance multi-label convolutional neural networks}.
\newblock In \emph{Proceedings of COLING 2016, the 26th International
  Conference on Computational Linguistics: Technical Papers}, pages 1471--1480.

\bibitem[{Keeler and Rumelhart(1992)}]{keeler1992self}
Jim Keeler and David~E Rumelhart. 1992.
\newblock A self-organizing integrated segmentation and recognition neural net.
\newblock In \emph{Advances in neural information processing systems}, pages
  496--503.

\bibitem[{Kingma and Ba(2014)}]{kingma2014adam}
Diederik~P Kingma and Jimmy Ba. 2014.
\newblock Adam: A method for stochastic optimization.
\newblock \emph{arXiv preprint arXiv:1412.6980}.

\bibitem[{Kotzias et~al.(2015)Kotzias, Denil, De~Freitas, and
  Smyth}]{kotzias2015group}
Dimitrios Kotzias, Misha Denil, Nando De~Freitas, and Padhraic Smyth. 2015.
\newblock From group to individual labels using deep features.
\newblock In \emph{Proceedings of the 21th ACM SIGKDD International Conference
  on Knowledge Discovery and Data Mining}, pages 597--606.

\bibitem[{Lei et~al.(2019)Lei, Yang, Yang, Zhao, Guo, and Liu}]{lei2019human}
Zeyang Lei, Yujiu Yang, Min Yang, Wei Zhao, Jun Guo, and Yi~Liu. 2019.
\newblock \href {https://doi.org/https://doi.org/10.1609/aaai.v33i01.33016650}
  {A human-like semantic cognition network for aspect-level sentiment
  classification}.
\newblock In \emph{Proceedings of the AAAI Conference on Artificial
  Intelligence}, volume~33, pages 6650--6657.

\bibitem[{Li et~al.(2017)Li, Guo, and Mei}]{li2017deep}
Cheng Li, Xiaoxiao Guo, and Qiaozhu Mei. 2017.
\newblock \href {https://doi.org/https://doi.org/10.1145/3018661.3018714} {Deep
  memory networks for attitude identification}.
\newblock In \emph{Proceedings of the Tenth ACM International Conference on Web
  Search and Data Mining}, pages 671--680.

\bibitem[{Li et~al.(2019)Li, Yin, Wei, Zhong, Luo, Xu, and Wu}]{Li2019AJM}
Yuncong Li, Cunxiang Yin, T.~Wei, Huiqiang Zhong, Jinchang Luo, Siqi Xu, and
  Xiaohui Wu. 2019.
\newblock A joint model for aspect-category sentiment analysis with
  contextualized aspect embedding.
\newblock \emph{ArXiv}, abs/1908.11017.

\bibitem[{Liang et~al.(2019)Liang, Meng, Zhang, Xu, Chen, and
  Zhou}]{liang2019novel}
Yunlong Liang, Fandong Meng, Jinchao Zhang, Jinan Xu, Yufeng Chen, and Jie
  Zhou. 2019.
\newblock \href {https://doi.org/https://doi.org/10.18653/v1/D19-1559} {A novel
  aspect-guided deep transition model for aspect based sentiment analysis}.
\newblock In \emph{Proceedings of the 2019 Conference on Empirical Methods in
  Natural Language Processing and the 9th International Joint Conference on
  Natural Language Processing (EMNLP-IJCNLP)}, pages 5572--5584.

\bibitem[{Liu(2012)}]{liu2012sentiment}
Bing Liu. 2012.
\newblock Sentiment analysis and opinion mining.
\newblock \emph{Synthesis lectures on human language technologies},
  5(1):1--167.

\bibitem[{Liu et~al.(2012)Liu, Wu, and Zhou}]{liu2012key}
Guoqing Liu, Jianxin Wu, and Zhi-Hua Zhou. 2012.
\newblock Key instance detection in multi-instance learning.

\bibitem[{Pang and Lee(2008)}]{pang2008opinion}
Bo~Pang and Lillian Lee. 2008.
\newblock Opinion mining and sentiment analysis.
\newblock \emph{Foundations and Trends{\textregistered} in Information
  Retrieval}, 2(1--2):1--135.

\bibitem[{Pappas and Popescu-Belis(2014)}]{pappas2014explaining}
Nikolaos Pappas and Andrei Popescu-Belis. 2014.
\newblock \href {https://doi.org/https://doi.org/10.3115/v1/D14-1052}
  {Explaining the stars: Weighted multiple-instance learning for aspect-based
  sentiment analysis}.
\newblock In \emph{Proceedings of the 2014 Conference on Empirical Methods In
  Natural Language Processing (EMNLP)}, pages 455--466.

\bibitem[{Paszke et~al.(2017)Paszke, Gross, Chintala, Chanan, Yang, DeVito,
  Lin, Desmaison, Antiga, and Lerer}]{paszke2017automatic}
Adam Paszke, Sam Gross, Soumith Chintala, Gregory Chanan, Edward Yang, Zachary
  DeVito, Zeming Lin, Alban Desmaison, Luca Antiga, and Adam Lerer. 2017.
\newblock Automatic differentiation in pytorch.

\bibitem[{Pennington et~al.(2014)Pennington, Socher, and
  Manning}]{pennington-etal-2014-glove}
Jeffrey Pennington, Richard Socher, and Christopher Manning. 2014.
\newblock \href {https://doi.org/10.3115/v1/D14-1162} {{G}love: Global vectors
  for word representation}.
\newblock In \emph{Proceedings of the 2014 Conference on Empirical Methods in
  Natural Language Processing ({EMNLP})}, pages 1532--1543, Doha, Qatar.
  Association for Computational Linguistics.

\bibitem[{Pontiki et~al.(2016)Pontiki, Galanis, Papageorgiou, Androutsopoulos,
  Manandhar, AL-Smadi, Al-Ayyoub, Zhao, Qin, De~Clercq, Hoste, Apidianaki,
  Tannier, Loukachevitch, Kotelnikov, Bel, Jim{\'e}nez-Zafra, and
  Eryi{\u{g}}it}]{pontiki-etal-2016-semeval}
Maria Pontiki, Dimitris Galanis, Haris Papageorgiou, Ion Androutsopoulos,
  Suresh Manandhar, Mohammad AL-Smadi, Mahmoud Al-Ayyoub, Yanyan Zhao, Bing
  Qin, Orph{\'e}e De~Clercq, V{\'e}ronique Hoste, Marianna Apidianaki, Xavier
  Tannier, Natalia Loukachevitch, Evgeniy Kotelnikov, Nuria Bel,
  Salud~Mar{\'\i}a Jim{\'e}nez-Zafra, and G{\"u}l{\c{s}}en Eryi{\u{g}}it. 2016.
\newblock \href {https://doi.org/10.18653/v1/S16-1002} {{S}em{E}val-2016 task
  5: Aspect based sentiment analysis}.
\newblock In \emph{Proceedings of the 10th International Workshop on Semantic
  Evaluation ({S}em{E}val-2016)}, pages 19--30, San Diego, California.
  Association for Computational Linguistics.

\bibitem[{Pontiki et~al.(2015)Pontiki, Galanis, Papageorgiou, Manandhar, and
  Androutsopoulos}]{pontiki-etal-2015-semeval}
Maria Pontiki, Dimitris Galanis, Haris Papageorgiou, Suresh Manandhar, and Ion
  Androutsopoulos. 2015.
\newblock \href {https://doi.org/10.18653/v1/S15-2082} {{S}em{E}val-2015 task
  12: Aspect based sentiment analysis}.
\newblock In \emph{Proceedings of the 9th International Workshop on Semantic
  Evaluation ({S}em{E}val 2015)}, pages 486--495, Denver, Colorado. Association
  for Computational Linguistics.

\bibitem[{Pontiki et~al.(2014)Pontiki, Galanis, Pavlopoulos, Papageorgiou,
  Androutsopoulos, and Manandhar}]{pontiki-etal-2014-semeval}
Maria Pontiki, Dimitris Galanis, John Pavlopoulos, Harris Papageorgiou, Ion
  Androutsopoulos, and Suresh Manandhar. 2014.
\newblock \href {https://doi.org/10.3115/v1/S14-2004} {{S}em{E}val-2014 task 4:
  Aspect based sentiment analysis}.
\newblock In \emph{Proceedings of the 8th International Workshop on Semantic
  Evaluation ({S}em{E}val 2014)}, pages 27--35, Dublin, Ireland. Association
  for Computational Linguistics.

\bibitem[{Ruder et~al.(2016)Ruder, Ghaffari, and
  Breslin}]{ruder-etal-2016-hierarchical}
Sebastian Ruder, Parsa Ghaffari, and John~G. Breslin. 2016.
\newblock \href {https://doi.org/10.18653/v1/D16-1103} {A hierarchical model of
  reviews for aspect-based sentiment analysis}.
\newblock In \emph{Proceedings of the 2016 Conference on Empirical Methods in
  Natural Language Processing}, pages 999--1005, Austin, Texas. Association for
  Computational Linguistics.

\bibitem[{Schmitt et~al.(2018)Schmitt, Steinheber, Schreiber, and
  Roth}]{schmitt2018joint}
Martin Schmitt, Simon Steinheber, Konrad Schreiber, and Benjamin Roth. 2018.
\newblock \href {https://doi.org/https://doi.org/10.18653/v1/D18-1139} {Joint
  aspect and polarity classification for aspect-based sentiment analysis with
  end-to-end neural networks}.
\newblock In \emph{Proceedings of the 2018 Conference on Empirical Methods in
  Natural Language Processing}, pages 1109--1114.

\bibitem[{Sun et~al.(2019)Sun, Huang, and Qiu}]{sun2019utilizing}
Chi Sun, Luyao Huang, and Xipeng Qiu. 2019.
\newblock \href {https://doi.org/https://doi.org/10.18653/v1/N19-1035}
  {Utilizing bert for aspect-based sentiment analysis via constructing
  auxiliary sentence}.
\newblock In \emph{Proceedings of the 2019 Conference of the North American
  Chapter of the Association for Computational Linguistics: Human Language
  Technologies, Volume 1 (Long and Short Papers)}, pages 380--385.

\bibitem[{Surdeanu et~al.(2012)Surdeanu, Tibshirani, Nallapati, and
  Manning}]{surdeanu2012multi}
Mihai Surdeanu, Julie Tibshirani, Ramesh Nallapati, and Christopher~D Manning.
  2012.
\newblock Multi-instance multi-label learning for relation extraction.
\newblock In \emph{Proceedings of the 2012 joint conference on empirical
  methods in natural language processing and computational natural language
  learning}, pages 455--465. Association for Computational Linguistics.

\bibitem[{Tay et~al.(2018)Tay, Tuan, and Hui}]{tay2018learning}
Yi~Tay, Luu~Anh Tuan, and Siu~Cheung Hui. 2018.
\newblock Learning to attend via word-aspect associative fusion for
  aspect-based sentiment analysis.
\newblock In \emph{Thirty-Second AAAI Conference on Artificial Intelligence}.

\bibitem[{Wang et~al.(2016)Wang, Huang, Zhu, and Zhao}]{wang2016attention}
Yequan Wang, Minlie Huang, Xiaoyan Zhu, and Li~Zhao. 2016.
\newblock \href {https://doi.org/https://doi.org/10.18653/v1/D16-1058}
  {Attention-based lstm for aspect-level sentiment classification}.
\newblock In \emph{Proceedings of the 2016 conference on empirical methods in
  natural language processing}, pages 606--615.

\bibitem[{Wang et~al.(2019)Wang, Sun, Huang, and Zhu}]{wang2019aspect}
Yequan Wang, Aixin Sun, Minlie Huang, and Xiaoyan Zhu. 2019.
\newblock \href {https://doi.org/https://doi.org/10.1145/3308558.3313750}
  {Aspect-level sentiment analysis using as-capsules}.
\newblock In \emph{The World Wide Web Conference}, pages 2033--2044.

\bibitem[{Wiegreffe and Pinter(2019)}]{wiegreffe2019attention}
Sarah Wiegreffe and Yuval Pinter. 2019.
\newblock \href {https://doi.org/https://doi.org/10.18653/v1/D19-1002}
  {Attention is not not explanation}.
\newblock In \emph{Proceedings of the 2019 Conference on Empirical Methods in
  Natural Language Processing and the 9th International Joint Conference on
  Natural Language Processing (EMNLP-IJCNLP)}, pages 11--20.

\bibitem[{Xing et~al.(2019)Xing, Liao, Song, Wang, Zhang, Wang, and
  Huang}]{xing2019earlier}
Bowen Xing, Lejian Liao, Dandan Song, Jingang Wang, Fuzheng Zhang, Zhongyuan
  Wang, and Heyan Huang. 2019.
\newblock Earlier attention? aspect-aware lstm for aspect sentiment analysis.
\newblock \emph{arXiv preprint arXiv:1905.07719}.

\bibitem[{Xue and Li(2018)}]{xue2018aspect}
Wei Xue and Tao Li. 2018.
\newblock \href {https://doi.org/https://doi.org/10.18653/v1/P18-1234} {Aspect
  based sentiment analysis with gated convolutional networks}.
\newblock In \emph{Proceedings of the 56th Annual Meeting of the Association
  for Computational Linguistics (Volume 1: Long Papers)}, pages 2514--2523.

\bibitem[{Yang et~al.(2016)Yang, Yang, Dyer, He, Smola, and
  Hovy}]{yang2016hierarchical}
Zichao Yang, Diyi Yang, Chris Dyer, Xiaodong He, Alex Smola, and Eduard Hovy.
  2016.
\newblock \href {https://doi.org/https://doi.org/10.18653/v1/N16-1174}
  {Hierarchical attention networks for document classification}.
\newblock In \emph{Proceedings of the 2016 conference of the North American
  chapter of the association for computational linguistics: human language
  technologies}, pages 1480--1489.

\bibitem[{Zhang and Zhou(2008)}]{zhang2008m3miml}
Min-Ling Zhang and Zhi-Hua Zhou. 2008.
\newblock M3miml: A maximum margin method for multi-instance multi-label
  learning.
\newblock In \emph{2008 Eighth IEEE International Conference on Data Mining},
  pages 688--697. IEEE.

\bibitem[{Zhou and Zhang(2006)}]{zhou2006multi}
Zhi-Hua Zhou and Min-Ling Zhang. 2006.
\newblock Multi-instance multi-label learning with application to scene
  classification.
\newblock In \emph{Proceedings of the 19th International Conference on Neural
  Information Processing Systems}, pages 1609--1616. MIT Press.

\bibitem[{Zhu et~al.(2019)Zhu, Chen, Zheng, and Qian}]{10.1145/3350487}
Peisong Zhu, Zhuang Chen, Haojie Zheng, and Tieyun Qian. 2019.
\newblock \href {https://doi.org/10.1145/3350487} {Aspect aware learning for
  aspect category sentiment analysis}.
\newblock \emph{ACM Trans. Knowl. Discov. Data}, 13(6).

\end{thebibliography}
\bibliographystyle{acl_natbib}

\end{document}